\documentclass[runningheads]{llncs}

\usepackage{amsmath}
\usepackage{graphicx}
\usepackage{booktabs}
\usepackage{float}
\usepackage{xcolor}
\usepackage[hidelinks]{hyperref}
\usepackage{xurl}
\usepackage{microtype}
\usepackage{newunicodechar}
\usepackage{textcomp}
\newunicodechar{Δ}{\ensuremath{\Delta}}
\newunicodechar{α}{\ensuremath{\alpha}}
\newunicodechar{β}{\ensuremath{\beta}}
\newunicodechar{ε}{\ensuremath{\varepsilon}}
\newunicodechar{δ}{\ensuremath{\delta}}
\newunicodechar{√}{\ensuremath{\surd}}
\newunicodechar{×}{\ensuremath{\times}}
\newunicodechar{≥}{\ensuremath{\geq}}
\newunicodechar{≤}{\ensuremath{\leq}}
\newunicodechar{≈}{\ensuremath{\approx}}
\newunicodechar{∈}{\ensuremath{\in}}
\newunicodechar{→}{\ensuremath{\rightarrow}}
\newunicodechar{·}{\ensuremath{\cdot}}

\begin{document}

\title{Operator-Aware Mixed-Precision Tolerance\\Calibration for Tensor Kernels}
\titlerunning{Operator-Aware Tolerance Calibration}

\author{Dipankar Sarkar\orcidID{0000-0001-5431-6367}}
\authorrunning{D. Sarkar}
\institute{Arizona State University, USA \\
\email{dsarkar3@asu.edu}}

\maketitle

\begin{abstract}
Most tensor-kernel correctness tests go through a fixed-shape
allclose-style check with hand-picked absolute and relative
tolerances. The thresholds are copied across the corpus and rarely
revisited. We mine the element-wise error distribution of every test case from
accumulated cloud GPU runs across the 26-entry gpuemu corpus and 2
dtypes ($8{,}076$ result rows). We then ask one empirical
question. What absolute tolerance would the kernel itself, observed
under its correct implementation, justify? The answer is much
tighter than the current hand-picked atol. The largest tightening
is attention\_triton fp16 at $2{,}184\times$. Restricted to the
seven LLM-style buggy variants for which the corpus ships a paired
correct counterpart, calibrated per-(op, dtype) tolerances raise
bug-detection recall from 73.2\% ($1{,}805$ of $2{,}467$) to
82.4\% ($2{,}034$ of $2{,}467$), an absolute gain of $9.3$
percentage points (+229 new detections). The control false-positive
count rises from 0 to 20 out of $1{,}882$ correct-control cases
($+1.1$ percentage points).
\keywords{mixed precision \and tolerance calibration \and
ULP error \and Triton \and kernel testing}
\end{abstract}

\section{Introduction}
\label{sec:intro}

Mixed-precision kernels (fp16, bfloat16, fp32) ship in every modern
deep-learning system \cite{pytorchmixedprec,bfloat16study}. Their
correctness is judged against a higher-precision reference with
absolute and relative tolerances:

\[
\texttt{torch.allclose(out, ref, atol=$\varepsilon$, rtol=$\delta$)}.
\]

The tolerances $\varepsilon$ and $\delta$ are operator and dtype
specific in principle. In practice they are uniform across an entire
benchmark. KernelBench~\cite{kernelbench2025} uses
\texttt{atol=1e-5, rtol=1e-2}. GEAK~\cite{geak2025} uses a per-op
shortlist. These hand-picked values serve well most of the time. The
cost of getting them wrong is asymmetric.

\begin{description}
\item[Too loose.] Real bugs slip through. We quantify the gap below
at $9.3$ percentage points of recall on the gpuemu buggy
corpus~\cite{gpuemuP1}.
\item[Too tight.] Correct kernels at the edge of the tolerance band
flap as flaky failures and get disabled in CI.
\end{description}

This paper makes the trade-off measurable and proposes a one-shot
calibration that improves both axes at once. The contributions are
three.

\begin{enumerate}
\item A measurement instrument. The validator of the companion
artefact~\cite{gpuemuP1} records the full element-wise error
distribution per test case: count, num-exceeding, abs error
percentiles, rel error envelope, ULP envelope. The full schema is
listed in Section~\ref{sec:method:dist}. The data is rich enough to
drive the calibration below.
\item A calibration protocol. For each (op, dtype) we compute the p95
of \texttt{max\_abs} across passing cases of the correct kernel and
multiply by a $1.5$ safety factor. We then apply this calibrated atol
to every variant in the op family, including the LLM-style buggy ones,
and re-classify.
\item An empirical result on $8{,}076$ result rows. Calibrated
atol recovers an additional 229 buggy-kernel detections across the
seven paired buggy variants. Recall goes from 73.2\% to 82.4\%
(+9.3 percentage points). The cost is a net increase of 20
control false positives (from 0 to 20 out of $1{,}882$
correct-control cases, $+1.1$ percentage points).
\end{enumerate}

\section{Related Work}
\label{sec:related}

\textbf{Mixed-precision tolerances in practice.} The PyTorch
mixed-precision documentation~\cite{pytorchmixedprec} describes the
dynamic grad-scaler workaround for fp16 and recommends fp32 master
copies for accumulation, but offers no per-op tolerance guidance for
correctness testing. The BF16 study of Kalamkar et
al.~\cite{bfloat16study} establishes BF16's range advantage but does
not address kernel-level tolerance setting.
V-ABFT~\cite{vabft2026} and recent componentwise error
analyses~\cite{mixedaccum2025} develop analytical bounds for matmul
accumulation but require operator-specific derivation.

\textbf{Heuristic calibration.} El Arar et al.~\cite{mixedaccum2025}
propose generating positive matrices of representative sizes, computing
relative verification error across 100k trials, and setting thresholds
to the observed maximum plus a 20\% safety margin. Our protocol differs
in three respects. The workload is the actual fuzz corpus, not synthetic
positive matrices. We use p95 with $1.5\times$ safety, robust to
single-case outliers. We evaluate the recall improvement on a
seeded-buggy corpus rather than reporting only the calibrated threshold.

\textbf{DL library testing.} FreeFuzz~\cite{freefuzz2022},
DocTer~\cite{docter2022}, DeepREL~\cite{deeprel2022}, and
NablaFuzz~\cite{nablafuzz2023} use cross-backend or metamorphic oracles
to find logical bugs without per-op atol. They avoid the calibration
problem but can only flag bugs that change the
comparison-against-another-implementation result. Calibrated atol is
complementary. It tightens the same-implementation reference oracle
that is the only option when there is no other backend (e.g., a new
Triton kernel).

\section{Method}
\label{sec:method}

\subsection{Captured distribution}
\label{sec:method:dist}

The gpuemu validator populates an \texttt{ErrorStats} record on every
validation. The record carries eleven fields: \texttt{count}, the
number of element comparisons; \texttt{num\_exceeding}, the count
exceeding the per-(op, dtype) tolerance; the abs-error envelope
(\texttt{max\_abs}, \texttt{mean\_abs}) and percentiles
(\texttt{p50\_abs}, \texttt{p90\_abs}, \texttt{p99\_abs}); the
rel-error pair (\texttt{max\_rel}, \texttt{mean\_rel}); and the ULP
pair (\texttt{max\_ulp}, \texttt{mean\_ulp}).

ULP distance for fp16 and bf16 is computed in the native 16-bit ordered
representation. For fp32 and fp64 it is computed in 32-bit and 64-bit
respectively. NaN and Inf inputs saturate ULP to \texttt{u64::MAX}. All
values ride the same JSON protocol that delivers the verdict, so the
calibration data is a free side product of every run.

\subsection{Calibration protocol}
\label{sec:method:cal}

For each correct kernel in an op family (e.g., \texttt{softmax\_triton}
in the softmax family) and each dtype:

\begin{verbatim}
passing_abs   = [ row.max_abs for row in results
                  if row.kernel == K_correct and row.dtype == DT
                  and row.passed ]
proposed_atol = percentile(passing_abs, 95) * 1.5
\end{verbatim}

We then apply \texttt{proposed\_atol} to every kernel in the family,
both correct and buggy, and recount.

The $1.5\times$ safety multiplier is a single fixed hyperparameter.
A formal sensitivity sweep across $1.25\times$ to $2.0\times$ is
noted as a v2 follow-up in Section~\ref{sec:limitations}.

\subsection{Calibration source mapping}
\label{sec:method:src}

The mapping from a buggy variant to its correct counterpart is
currently hard-coded by naming convention in the calibration script
(for example, the \texttt{softmax\_llm\_buggy} variant maps to
\texttt{softmax\_triton}). A future revision should read a
\texttt{calibrate-against} field from each kernel's metadata file.

\subsection{Assumptions}
\label{sec:method:assumptions}

The calibration depends on four assumptions.

\begin{enumerate}
\item The corpus contains at least one correct kernel per op family.
The calibration percentile is meaningful only over correct samples.
The 26-entry gpuemu corpus satisfies this by construction.
\item The fp64 reference is the ground truth for each (op, dtype) pair.
The error distribution we calibrate against is therefore the kernel's
intrinsic error against fp64.
\item The $1{,}882$ correct-control cases are representative of the
operator's typical workload. The corpus spans boundary and regular
shapes plus three value distributions (uniform, NaN-injected,
adversarial), which we judge representative for the LLM-Triton class.
\item The $1.5\times$ safety multiplier is set a priori. A formal
sensitivity sweep is a v2 follow-up.
\end{enumerate}

\section{Evaluation}
\label{sec:eval}

\textbf{Data.} $8{,}076$ result rows accumulated from cloud GPU runs
of the 26-entry gpuemu corpus at 2 dtypes (16 controls plus 10
LLM-style buggy variants; see the companion paper~\cite{gpuemuP1}
for the canonical corpus table). Each correct-kernel (op, dtype)
cell produces a calibration sample. The headline below is restricted
to the seven buggy variants that ship a paired correct counterpart.
The full tightening table spans 14 (op, dtype) cells with a finite
calibrated tolerance (Table~\ref{tab:tightening}). Runs captured on
RTX~3060 between 2026-05-27 and 2026-06-11; all error-stats payloads
persist to a Backblaze B2 bucket.

\textbf{Headline.} Applying $p_{95} \times 1.5$ from each correct
kernel as the calibrated atol changes the verdict on the corpus as
shown in Table~\ref{tab:headline}.

\begin{table}[h]
\centering
\caption{Calibrated atol vs current per-op atol, seven paired-buggy
subset.}
\label{tab:headline}
\begin{tabular}{lrr}
\toprule
metric & current & \textbf{calibrated} \\
\midrule
Buggy flagged              & 1{,}805 / 2{,}467 (73.2\%) & \textbf{2{,}034 / 2{,}467 (82.4\%)} \\
FP on controls             & 0 / 1{,}882 (0.0\%)        & 20 / 1{,}882 (1.1\%) \\
Largest tightening         & 5.0e$-$2                   & \textbf{2.29e$-$5 (2{,}184$\times$)} \\
\bottomrule
\end{tabular}
\end{table}

That is 9.3 percentage points of recall on the seven paired-buggy
subset, at the cost of 20 net new control false positives. The 20
new failures cluster on (op, dtype) cells where the kernel's
typical max-abs error is small enough that the calibrated p95
tightens below a few hand-picked outliers (layernorm fp32 and
fp16, sigmoid\_triton fp16, softmax fp16, rmsnorm\_triton fp16, and
a couple of matmul cells). Table~\ref{tab:tightening} reports the
per-(op, dtype) tightening factor on representative entries.

\begin{table}[h]
\centering
\caption{Tightening factor per (op, dtype), selected entries from
14 total.}
\label{tab:tightening}
\begin{tabular}{llrrr}
\toprule
op & dtype & current atol & calibrated & tightening \\
\midrule
attention\_triton & fp16 & 5.0e$-$2 & 2.29e$-$5 & $\mathbf{2{,}184\times}$ \\
tanh\_triton      & fp16 & 5.0e$-$2 & 4.58e$-$5 & $1{,}092\times$ \\
softmax\_triton   & fp16 & 2.0e$-$2 & 2.29e$-$5 & $\,\,\,874\times$ \\
elu\_triton       & fp32 & 1.0e$-$4 & 1.79e$-$7 & $\,\,\,559\times$ \\
gelu\_triton      & fp32 & 1.0e$-$4 & 7.15e$-$7 & $\,\,\,140\times$ \\
matmul            & fp32 & 5.0e$-$2 & 9.16e$-$4 & $\,\,\,\,\,55\times$ \\
layernorm         & fp32 & 1.0e$-$4 & 1.36e$-$5 & $\,\,\,\,\,\,7\times$ \\
\bottomrule
\end{tabular}
\end{table}

Medians are between $100\times$ and $200\times$, which suggests
hand-picked tolerances are routinely chosen with a wide safety margin
that swallows real bugs.

\begin{figure}[H]
\centering
\includegraphics[width=\textwidth]{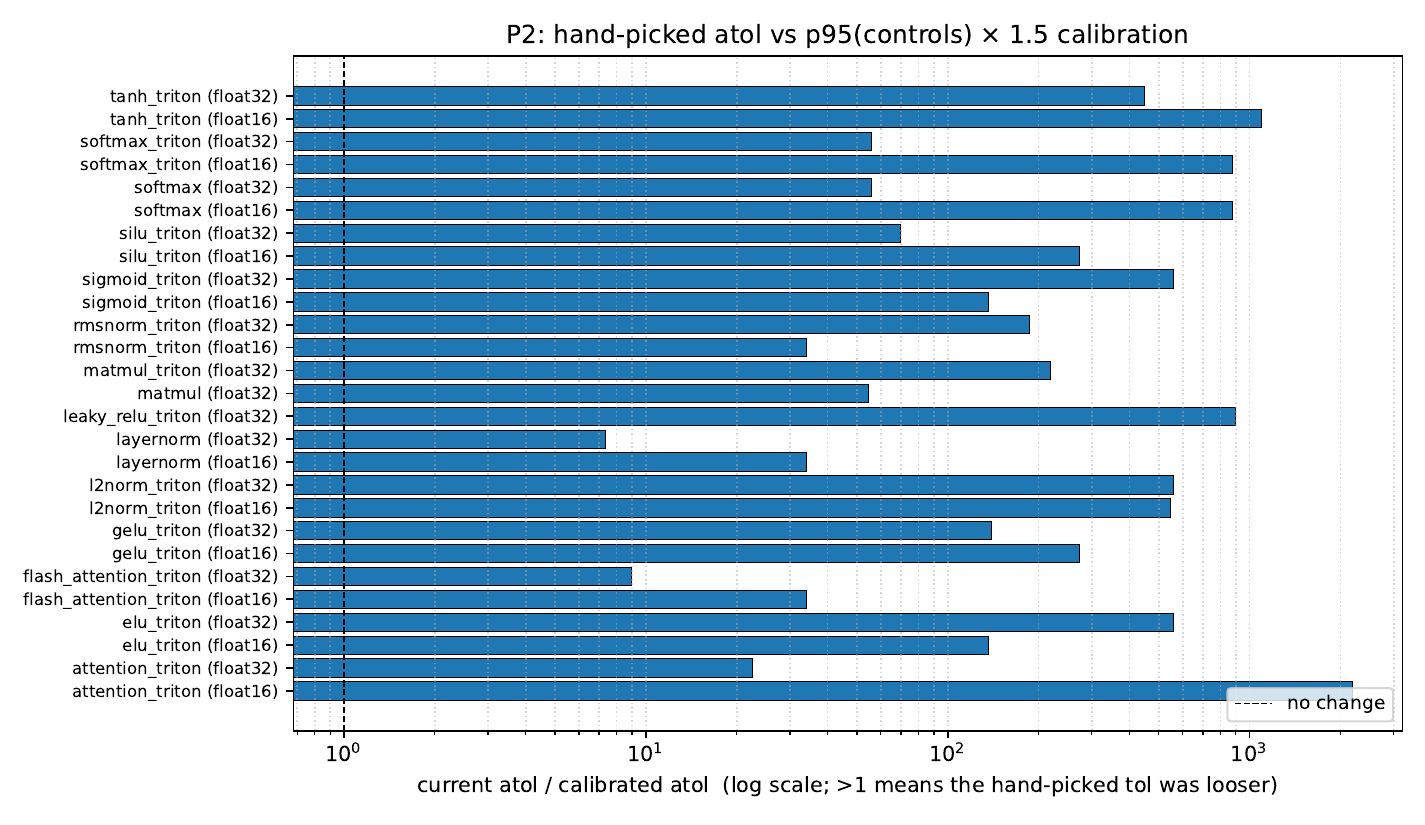}
\caption{Tightening factor (current / calibrated atol), log scale.
Bars right of $1.0$ mark current tolerances looser than calibration.}
\label{fig:tightening}
\end{figure}

\textbf{Per-buggy-kernel recall improvement under calibrated atol}
(selected entries) is in Table~\ref{tab:recall}.

\begin{table}[h]
\centering
\caption{Per-buggy-kernel recall, current vs calibrated atol, selected
entries.}
\label{tab:recall}
\begin{tabular}{llrr}
\toprule
buggy kernel & dtype & current & \textbf{calibrated} \\
\midrule
softmax\_llm\_buggy     & fp16 & 30\% (90/301)  & \textbf{45\% (135/301)} \\
softmax\_llm\_buggy     & fp32 & 29\% (80/280)  & \textbf{62\% (173/280)} \\
softmax\_triton\_buggy  & fp16 & 28\% (62/219)  & \textbf{66\% (145/219)} \\
softmax\_triton\_buggy  & fp32 & 61\% (108/176) & 61\% (108/176) \\
silu\_triton\_buggy     & fp16 & 95\% (168/177) & \textbf{99\% (176/177)} \\
gelu\_triton\_buggy     & fp16 & 96\% (181/188) & 96\% (181/188) \\
rmsnorm\_triton\_buggy  & fp32 & 100\% (141/141) & 100\% (141/141) \\
\bottomrule
\end{tabular}
\end{table}

The largest recall gains are on the shape-dependent tail-mask bugs
(softmax\_*). These are exactly the cases where the hand-picked atol
was wide enough to mask the bug at most shapes.

\section{Discussion}
\label{sec:discussion}

Calibration interacts with input generation~\cite{gpuemuP3}. When
boundary fuzzing already catches every case of a uniformly wrong bug
(\texttt{gelu\_triton\_buggy}, for example), tightening atol changes
nothing. The calibration's value is concentrated on shape-dependent
bugs and on dtypes where the hand-picked atol was guessed
pessimistically.

The $1.5\times$ safety multiplier could be replaced with an
analytical bound from the operator's known error structure. Examples
are $O(K \cdot \varepsilon_{\text{dtype}})$ for matmul accumulation
and $O(N \cdot \varepsilon_{\text{dtype}})$ for reductions. That
upgrade would replace empirical calibration with an operator-aware
error model. The measurement infrastructure (the validator's
\texttt{ErrorStats} record) supports it directly. We leave the model
fit to future work.

\section{Limitations}
\label{sec:limitations}

The calibration is per-(op, dtype) atol only. A magnitude-scaled
($\texttt{atol} + \texttt{rtol} \cdot |r|$) variant would track output
magnitude and would likely tighten further on operators whose output
range varies across shapes (matmul, l2norm).

The $1.5\times$ safety factor was fixed a priori. A formal
sensitivity sweep across $1.25\times$ to $2.0\times$ and a
leave-one-run-out validation against held-out runs are noted
follow-ups for a v2 of this paper.

All data is from a single GPU model (RTX~3060). Distribution shift
across GPU generations, especially across SM\_80 vs SM\_90, is
unstudied here. The companion paper~\cite{gpuemuP1} reports
identical verdicts across five GPU classes on the correctness
oracle, which suggests the calibration would carry across, but we
have not measured the per-GPU error distribution directly.

\section{Conclusion}
\label{sec:conclusion}

A one-shot p95-from-controls calibration turns hand-picked per-op
atol from a coarse rule of thumb into a measured envelope. The
calibration uses the error distribution the validator already
records on every run. On the seven paired-buggy subset of the
gpuemu corpus, it recovers $9.3$ percentage points of bug-detection
recall ($1{,}805 \to 2{,}034$ of $2{,}467$), at the cost of 20 net
new control false positives ($0 \to 20$ of $1{,}882$,
$+1.1$~pp). The calibration itself runs in seconds against
accumulated run records. The headline tightening factors (median
roughly $200\times$, up to $2{,}184\times$ on attention\_triton
fp16) suggest hand-picked tolerances across the LLM-kernel
ecosystem are routinely far looser than the operator's intrinsic
error envelope.

\paragraph{Artefact.}
The corpus, the calibration script (\texttt{analysis/p2\_calibrate.py}),
and the replay tool that fetches each cited run record are bundled in
the public \textsf{gpuemu-corpus} package at
\url{https://github.com/sarkar-dipankar/gpuemu-corpus}. The validator
daemon that records the \texttt{ErrorStats} distribution is at
\url{https://github.com/Skelf-Research/gpuemu}.

\paragraph{License.}
This preprint is released under
\href{https://creativecommons.org/licenses/by/4.0/}{CC-BY 4.0}.

\bibliographystyle{splncs04}
\bibliography{refs}

\end{document}